\documentclass[10pt,twocolumn,letterpaper]{article}

\usepackage[pagenumbers]{cvpr} 

\usepackage{graphicx}
\usepackage{amsmath}
\usepackage{amssymb}
\usepackage{booktabs}

\usepackage[pagebackref,breaklinks,colorlinks]{hyperref}

\usepackage[capitalize]{cleveref}
\crefname{section}{Sec.}{Secs.}
\Crefname{section}{Section}{Sections}
\Crefname{table}{Table}{Tables}
\crefname{table}{Tab.}{Tabs.}

\def\cvprPaperID{6982} 
\def\confName{CVPR}
\def\confYear{2022}

\usepackage{bibentry}
\usepackage{amsmath}
\usepackage{multirow}
\begin{document}

\title{E2EC: An End-to-End Contour-based Method for High-Quality High-Speed Instance Segmentation}

\author{
{
Tao Zhang$^{*}$, ~Shiqing Wei\thanks{ : Equally contributed.}, ~Shunping Ji\thanks{ : Corresponding author.}}\\
{
Wuhan University, China}\\
\small \tt \{zhang\_tao,  wei\_sq, jishunping\}@whu.edu.cn 
}
\maketitle

\begin{abstract}
Contour-based instance segmentation methods have developed rapidly recently but feature rough and handcrafted front-end contour initialization, which restricts the model performance, and an empirical and fixed backend predicted-label vertex pairing, which contributes to the learning difficulty. In this paper, we introduce a novel contour-based method, named E2EC, for high-quality instance segmentation. Firstly, E2EC applies a novel learnable contour initialization architecture instead of handcrafted contour initialization. This consists of a contour initialization module for constructing more explicit learning goals and a global contour deformation module for taking advantage of all of the vertices' features better. Secondly, we propose a novel label sampling scheme, named multi-direction alignment, to reduce the learning difficulty. Thirdly, to improve the quality of the boundary details, we dynamically match the most appropriate predicted-ground truth vertex pairs and propose the corresponding loss function named dynamic matching loss. The experiments showed that E2EC can achieve a state-of-the-art performance on the KITTI INStance (KINS) dataset, the Semantic Boundaries Dataset (SBD), the Cityscapes and the COCO dataset. E2EC is also efficient for use in real-time applications, with an inference speed of 36 fps for 512×512 images on an NVIDIA A6000 GPU. Code will be released at \href{https://github.com/zhang-tao-whu/e2ec}{https://github.com/zhang-tao-whu/e2ec}.
\end{abstract}\vspace{-5mm}
\section{Introduction}
\label{sec:intro}
Instance segmentation is a fundamental computer vision task and the cornerstone of many downstream computer vision applications, such as autonomous driving and robotic grasping. The classic instance segmentation methods are based on a two-stage pipeline, where the bounding boxes (bboxes) of the instances are first generated, and then pixel-wise segmentation is performed within the bboxes. Typical examples are methods such as Mask R-CNN \cite{maskrcnn} and PANet \cite{panet}. These methods can achieve an excellent accuracy, but are inefficient, which restricts their application in real-time tasks. With the rapid development of one-stage detectors \cite{centernet,fcos}, many one-stage mask-based instance segmentation methods have now been proposed, such as YOLACT \cite{yolact}, BlendMask \cite{blendmask}, TensorMask \cite{tensormask}, and CenterMask \cite{centermask}. However, these one-stage methods consume a lot of storage, require costly post-processing, and hardly perform in real time. The quality of the instance boundary prediction is also unsatisfactory as these methods usually use limited feature information (for example, Mask R-CNN only segments instances in a 28×28 feature map).
\begin{figure}[t]
\begin{minipage}[c]{0.33\linewidth}
\includegraphics[width=1.0\linewidth]{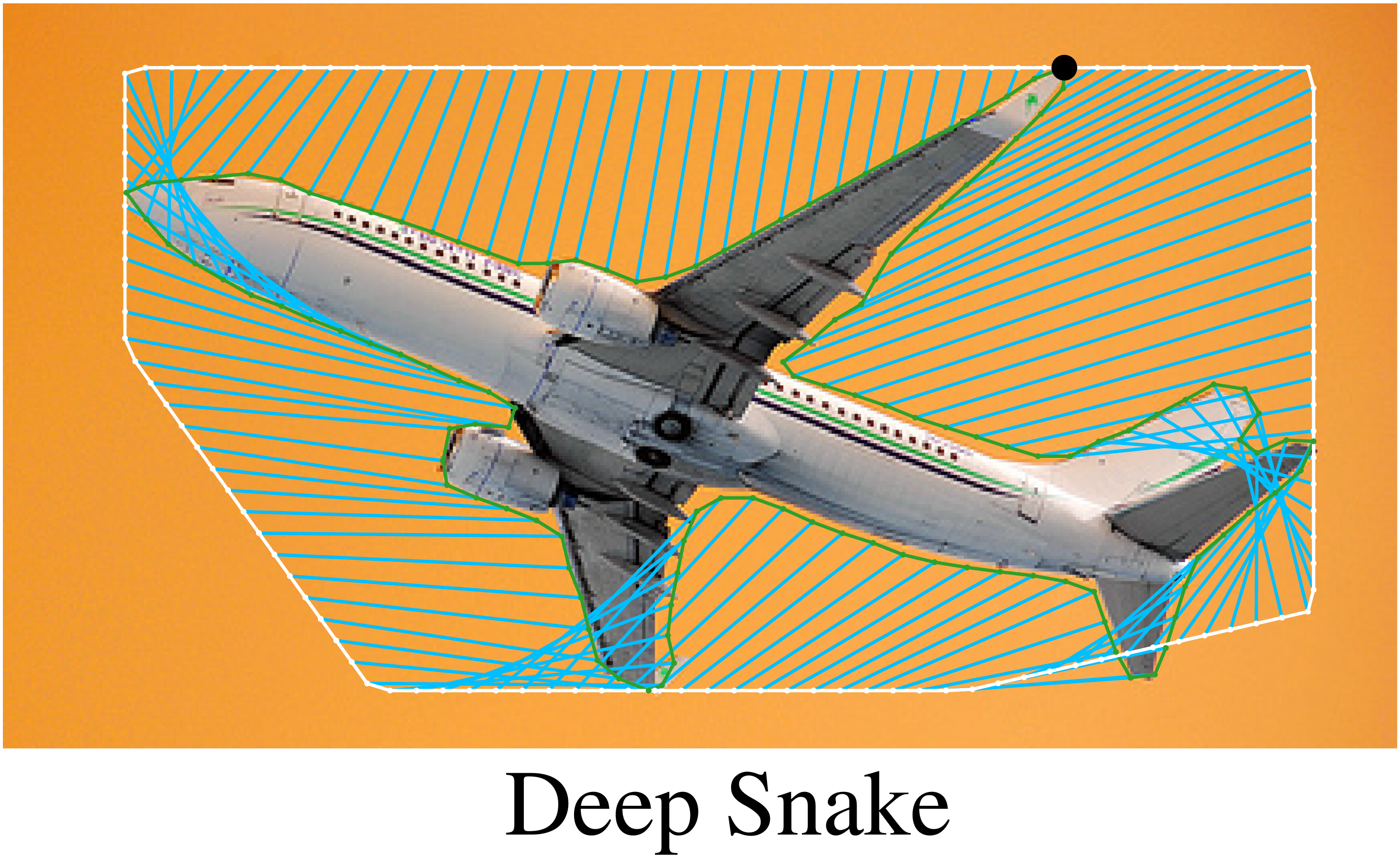}
\end{minipage}\hfill
\begin{minipage}[c]{0.33\linewidth}
\includegraphics[width=1.0\linewidth]{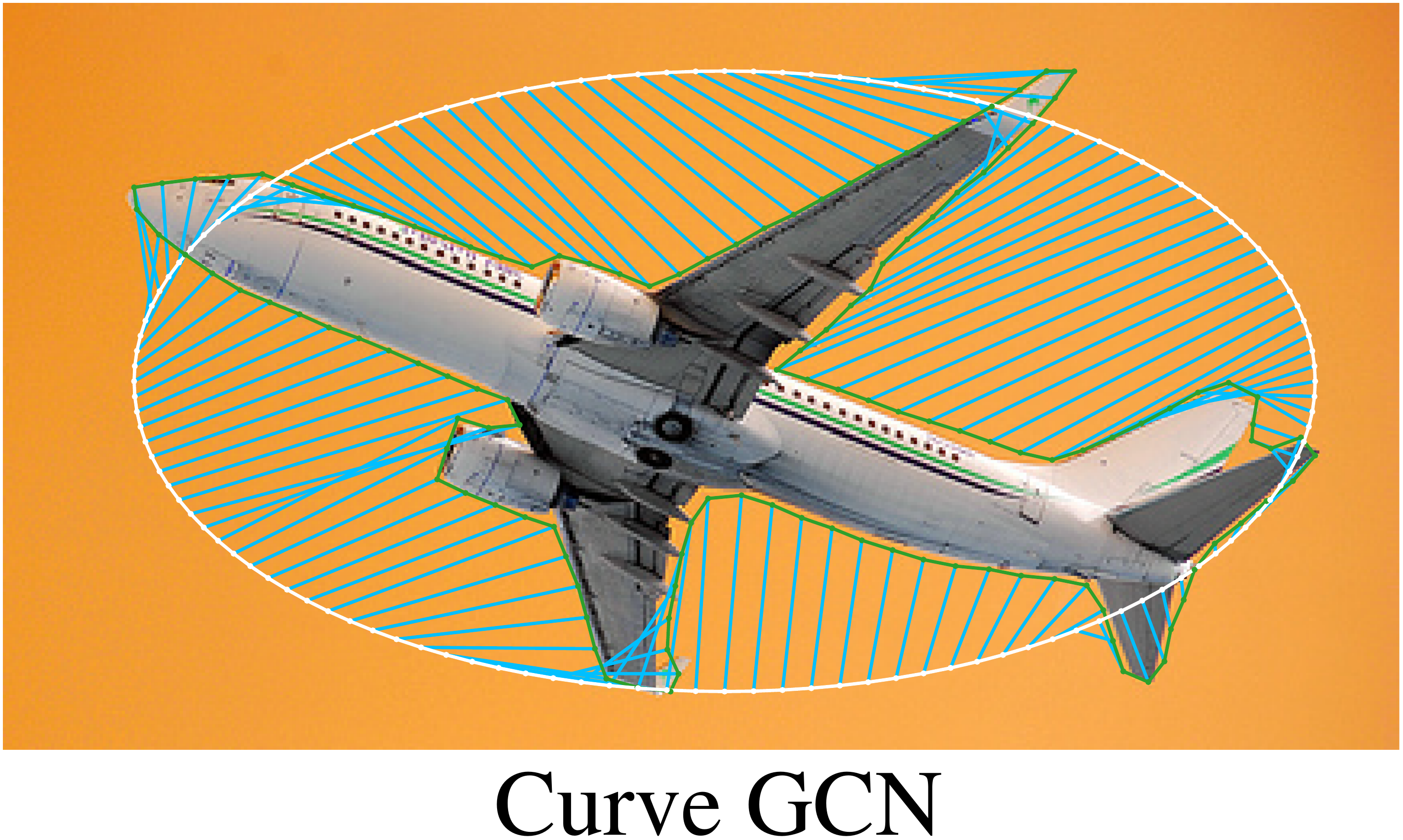}
\end{minipage}\hfill
\begin{minipage}[c]{0.33\linewidth}
\includegraphics[width=1.0\linewidth]{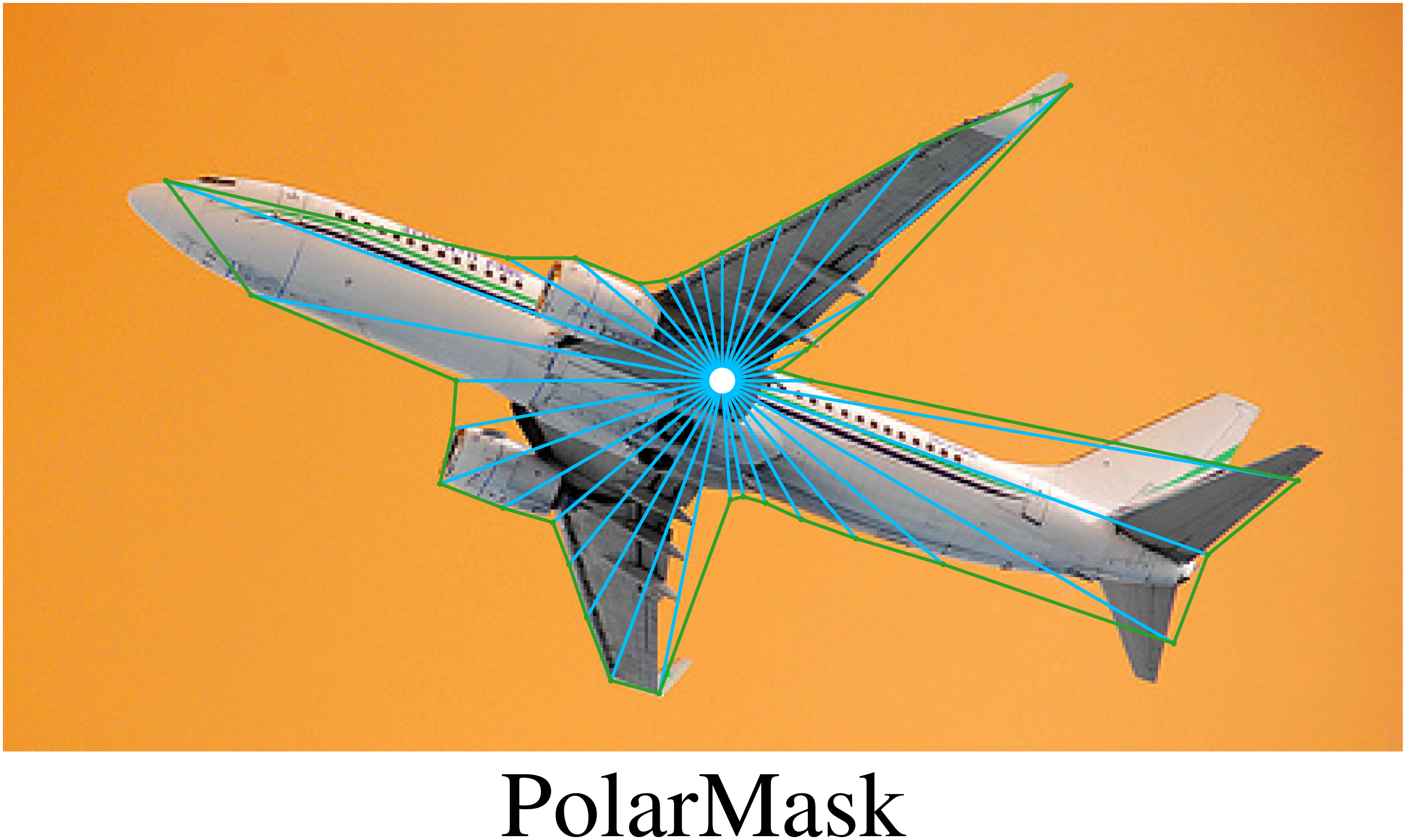}
\end{minipage}\hfill
\begin{minipage}[c]{0.33\linewidth}
\includegraphics[width=1.0\linewidth]{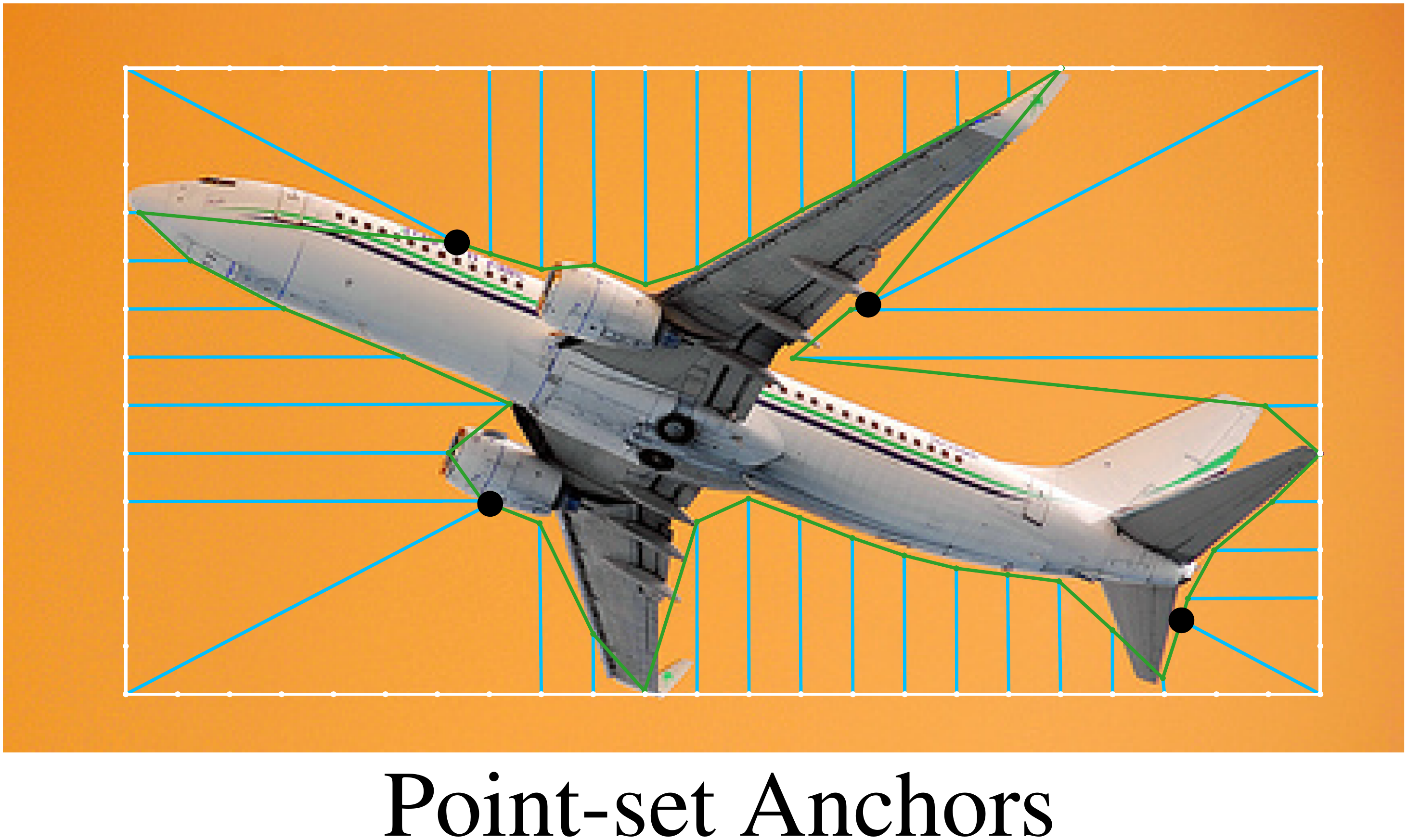}
\end{minipage}\hfill
\begin{minipage}[c]{0.33\linewidth}
\includegraphics[width=1.0\linewidth]{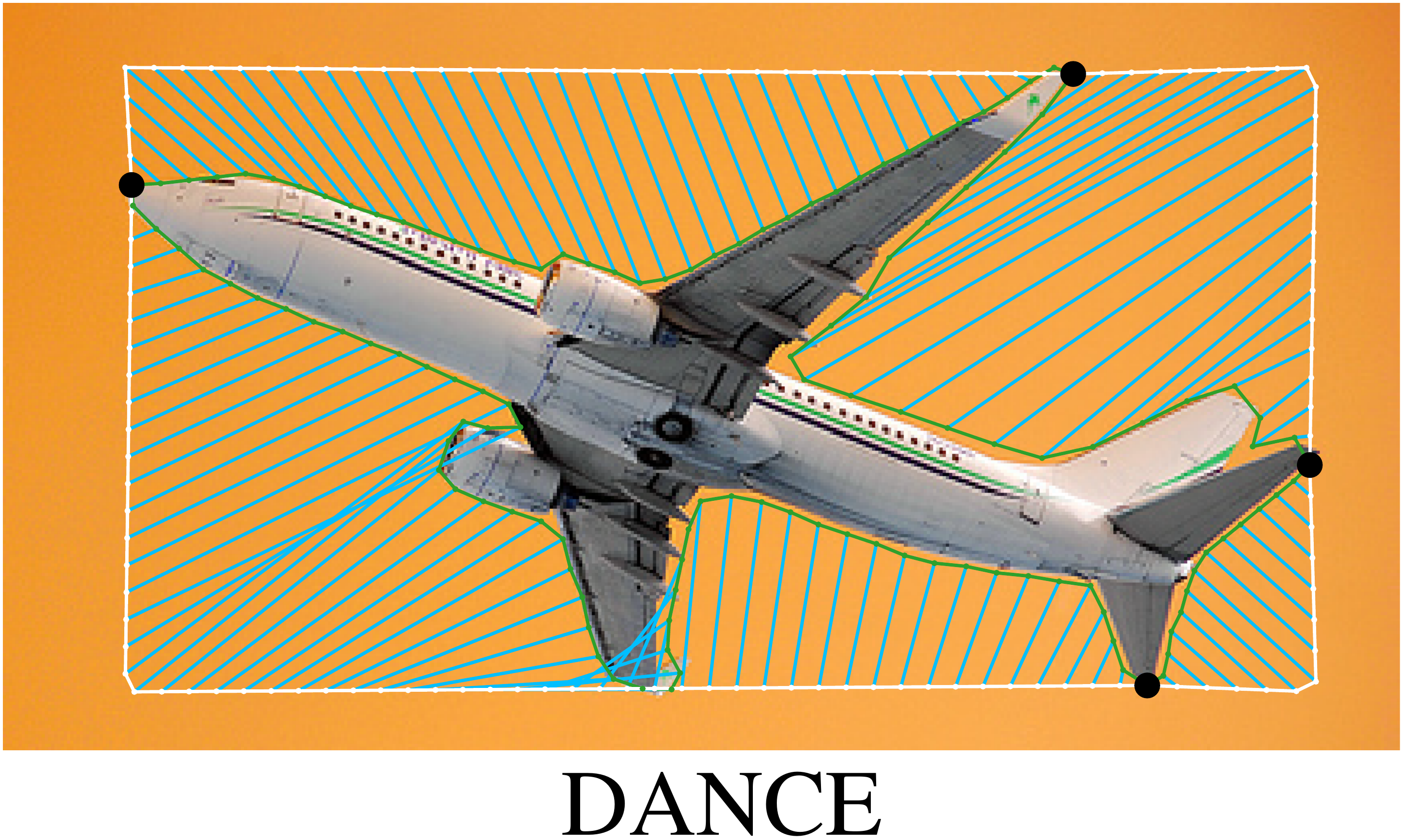}
\end{minipage}\hfill
\begin{minipage}[c]{0.33\linewidth}
\includegraphics[width=1.0\linewidth]{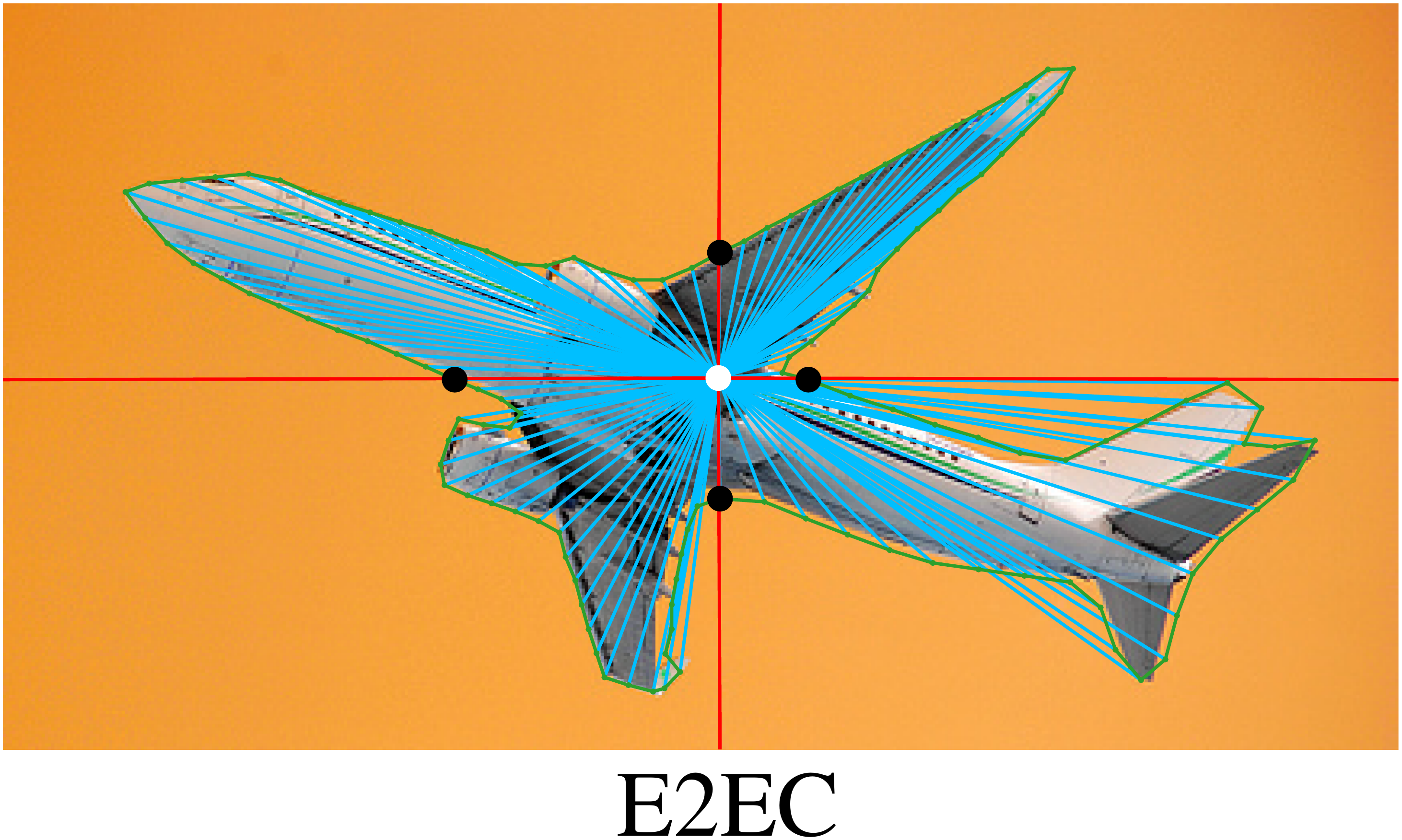}
\end{minipage}\hfill
\vspace{-2mm}
\caption{\textbf{The ideal deformation paths of several contour-based methods.} White boundaries and points are the initial contours, blue lines are the deformation paths, and black points are the alignment points.}\vspace{-8mm}
\label{fig:deform path}
\end{figure}
The contour-based methods have recently received renewed attention and have shown great potential. Examples of such methods are Curve GCN \cite{curvegcn}, Deep Snake \cite{deepsnake}, Point-Set Anchors \cite{pointset}, DANCE \cite{dance}, PolarMask \cite{polarmask}, and LSNet \cite{lsnet}. The contour-based methods treat the instance segmentation as a regression task, i.e., regressing the vertex coordinates of a contour represented by a series of discrete vertices. A contour composed of $N$ (e.g., $N=128$) vertices is sufficient to describe most of the instances well \cite{deepsnake}. Compared with the mask-based methods, which require intensive processing of each pixel, the contour-based methods are simpler and require less calculation. The contour-based methods can also directly obtain the boundaries of the instances, without any complicated post-processing.
 
However, the existing contour-based methods still have many obvious shortcomings. First, all of the existing multi-stage methods adopt a manually designed shape for the initial contour. As shown in Figure \ref{fig:deform path}, the difference between the manually designed initial contour and the ground-truth instance boundary can lead to many unreasonable deformation paths (the route from initial to ground-truth vertex) and huge training difficulty. It is also impossible to sample the manually designed initial contour to achieve a uniform angle and uniform vertex spacing at the same time. The Point-Set Anchors and DANCE methods attempt to address this problem by changing the intuitive vertex pairing method \cite{pointset, dance}, but the results are not satisfactory.

Second, local or limited information is popularly applied in contour adjustment. For example, the one-stage PolarMask \cite{polarmask} and LSNet \cite{lsnet} methods directly regress the coordinates of the contour vertices based on only the limited features at the instance center, resulting in the loss of the predicted contour details. The multi-stage methods iteratively adjust the initial contour based on the features of the contour vertices to obtain a more refined segmentation result. However, Curve GCN and Deep Snake utilize a local information aggregation mechanism that propagates the features of the local adjacent contour vertices to refine the contour, which can fail in correcting large prediction errors. Moreover, the local aggregation has to be inefficiently repeated to access global information. Instead, we propose a global contour deformation method based on the features of all the contour vertices.

Third, the pairing of the ground-truth and predicted vertices in the current contour-based methods is fixed, regardless of the continuous position adjustment of a predicted vertex (e.g., it is already on the ground-truth boundary or close to another ground-truth vertex, but far from the given one). Hence, the pre-fixed vertex pairing is not optimal, and can result in a slower convergence speed, and even wrong predictions.

In this paper, we propose a multi-stage and highly efficient end-to-end contour-based instance segmentation model named E2EC, which can completely overcome these shortcomings. E2EC incorporates three novel components: 1) a learnable contour initialization architecture; 2) multi-direction alignment (MDA); and 3) a dynamic matching loss (DML) function.

E2EC replaces the manually designed initial contour with a learnable contour initialization architecture, which handles the first and second problem. This architecture contains two novel modules: 1) a contour initialization module; and 2) a global contour deformation module. The contour initialization module directly regresses the complete initial contour based on the center point features, which differs from regressing lengths along given fixed rays \cite{polarmask}. The global contour deformation module then refines the initial contour based on all of the features of the initial contour vertices and center point instead of using features of local vertices. As shown in Figure \ref{fig:deform path}, the learnable initial contour architecture does not rely on a manually designed initial contour (e.g., the ellipse of Curve GCN or the octagon of Deep Snake), and directly deforms from the midpoint of the object instance to the contour with more reasonable paths.
\begin{figure}[t]
  \centering
   \includegraphics[width=1.0\linewidth]{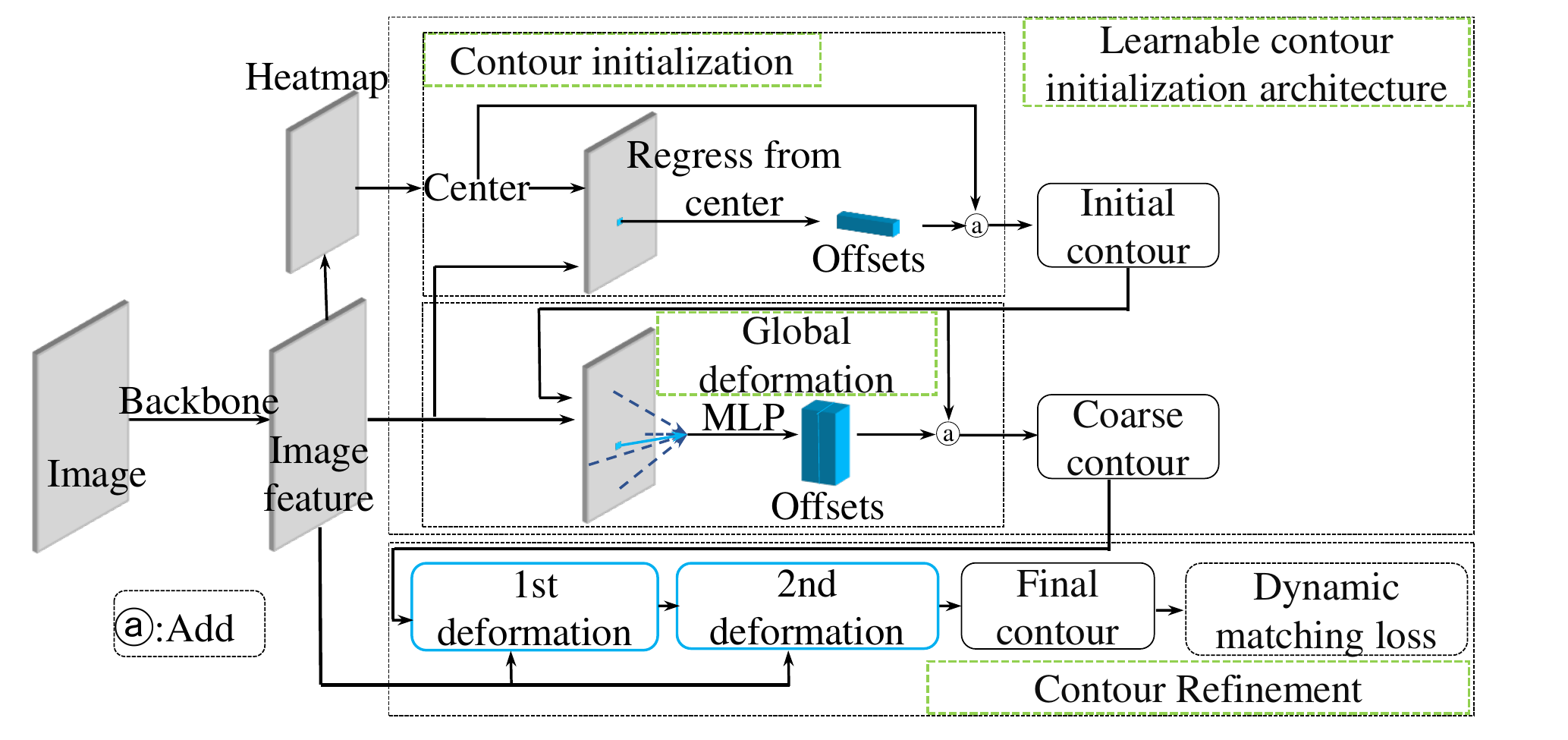}\vspace{-3mm}
   \caption{\textbf{Overview of E2EC.} E2EC consists of a learnable contour initialization architecture including a contour initialization and a global deformation module that produces the coarse contour, and a contour refinement module that produces the final contour with the supervision of DML.}
   \vspace{-11mm}
   \label{fig:pipeline}
\end{figure}

The difficulty in predicted-label vertex pairing roots in the fact that no simple differentiable calculation can measure the distance between the predicted and ground-truth boundaries. To address the third problem, on the one hand, we propose multi-direction alignment (MDA), which fixes the directions of the selected multiple contour vertices with respect to the center point (the black points in Figure \ref{fig:deform path} (E2EC)), and then uniformly samples between the fixed vertices to generate ground-truth vertices. MDA appropriately restricts the possible vertex pairing and deformation paths, and greatly reduces the difficulty of learning while ensuring the upper bound of the performance. The combination of the learnable initial contour architecture and MDA eliminates the unreasonable deformation paths that commonly exist in the current contour-based methods.

On the other hand, we propose a matching strategy that dynamically matches the predicted vertices and the most appropriate label vertices instead of fixed pairing, and the corresponding dynamic matching loss (DML) function. DML eliminates the problems of an over-smooth boundary and poor fitting of the inflection points in the contour-based methods, and greatly improves the quality of the predicted boundary details. 

In the experiments conducted in this study, E2EC exhibited a state-of-the-art performance on the KITTI INStance (KINS) dataset \cite{kins}, the Semantic Boundaries Dataset (SBD) \cite{sbd} and the Cityscapes \cite{cityscapes} dataset. For 512×512 images, E2EC  achieved a 36 fps inference speed on an NVIDIA A6000 GPU. If the iterative deformation module is disabled, E2EC can reach a speed of 50 fps, with an accuracy comparable to that of Deep Snake.
\section{Related Work}
\label{sec:related work}

\textbf{Mask-based instance segmentation methods.} The classic mask-based instance segmentation methods, such as Mask R-CNN \cite{maskrcnn} and PANet \cite{panet}, include a bbox extraction stage and a mask segmentation stage. These methods can achieve a good performance, but with a very slow speed. In recent years, one-stage methods such as CenterMask \cite{centermask}, YOLACT \cite{yolact}, SOLO \cite{solo}, and BlendMask \cite{blendmask} that follow the above process have developed rapidly, and have greatly improved in speed. However, dense pixel-wise classification requires a huge amount of calculation. Although these methods try to sacrifice performance and perform segmentation on the down-sampled feature maps, to reduce the amount of calculation, they still cannot meet the requirement of real-time performance. The methods proposed in \cite{ssap, spatial} follow another pipeline, where they first perform semantic segmentation and then cluster the pixels to generate instances. However, these methods require complex post-processing and cannot be applied to amodal instance segmentation tasks.
\begin{figure}[t]
\begin{minipage}[c]{0.5\linewidth}\vspace{-3mm}
\includegraphics[width=0.95\linewidth]{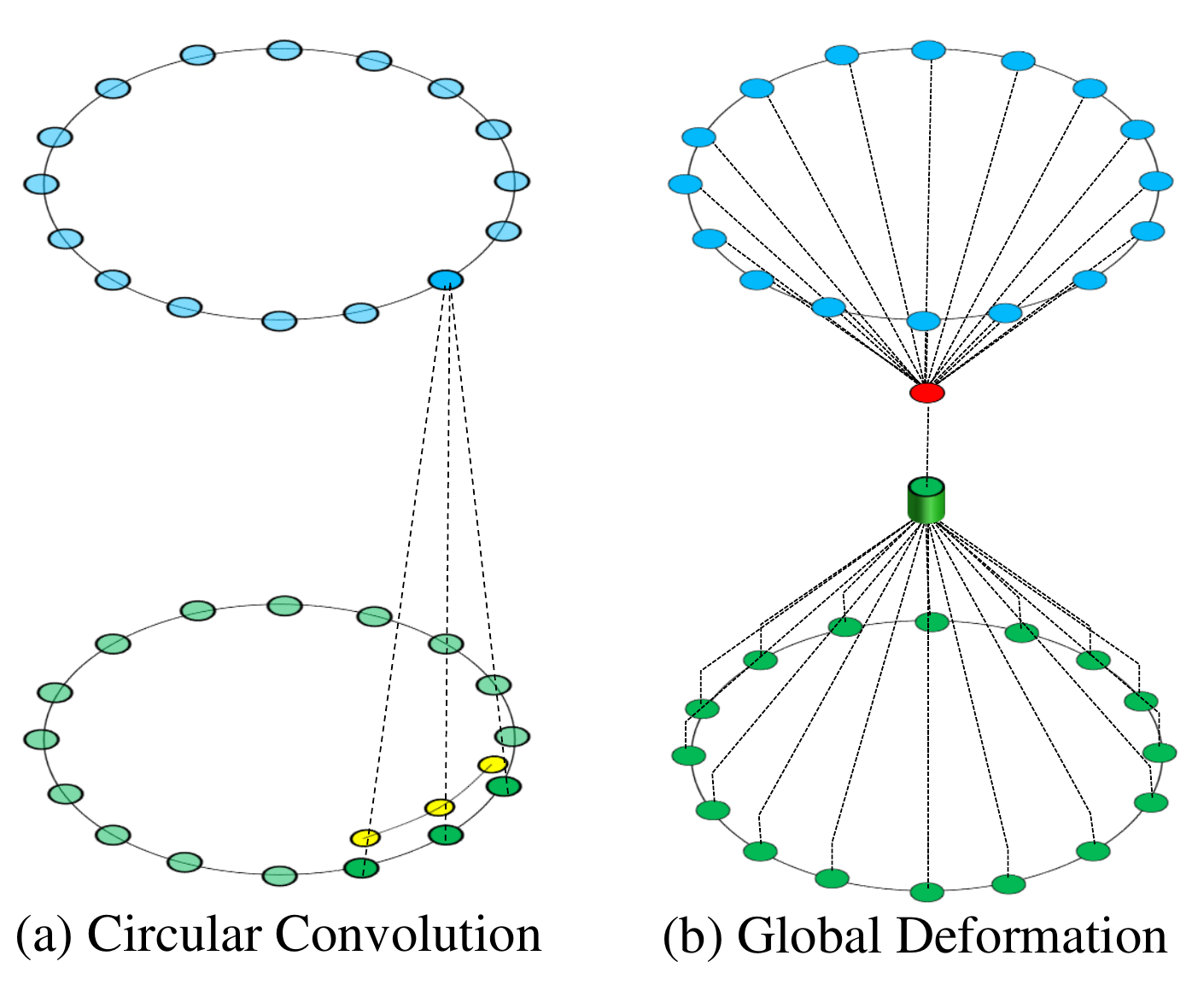}
\end{minipage}\hfill
\begin{minipage}[c]{0.5\linewidth}
   \caption{\textbf{Global deformation (b) vs. circular convolution \cite{deepsnake} (a).} The green points represent the features of the contour vertices, the yellow points represent the local kernel function of the circular convolution, and the blue points represent the offsets of the contour vertices, and red is MLP.}
   \label{fig:global deformation}
\end{minipage}\vspace{-8mm}
\end{figure}

\begin{figure}[t]
\begin{minipage}[c]{0.25\linewidth}
\includegraphics[width=1.0\linewidth]{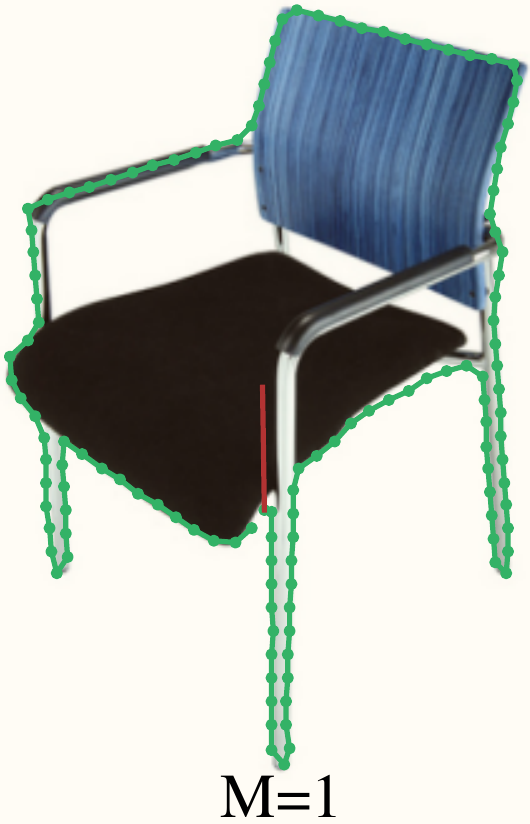}
\end{minipage}\hfill
\begin{minipage}[c]{0.25\linewidth}
\includegraphics[width=1.0\linewidth]{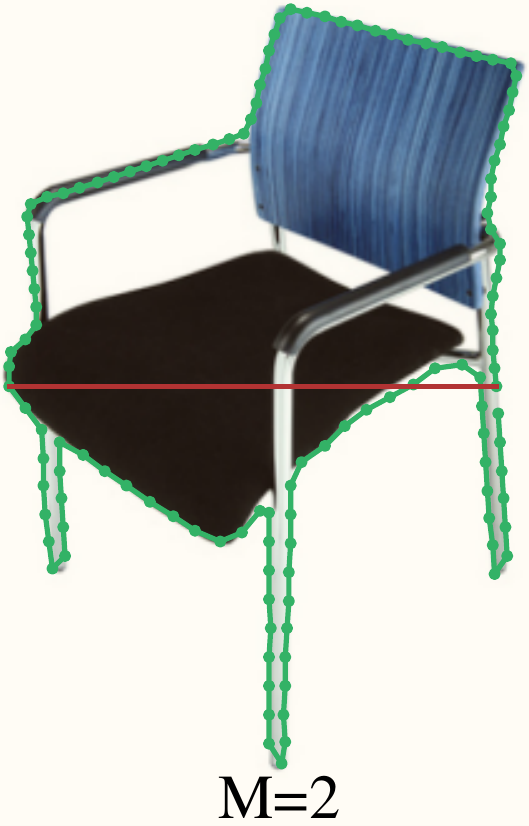}
\end{minipage}\hfill
\begin{minipage}[c]{0.25\linewidth}
\includegraphics[width=1.0\linewidth]{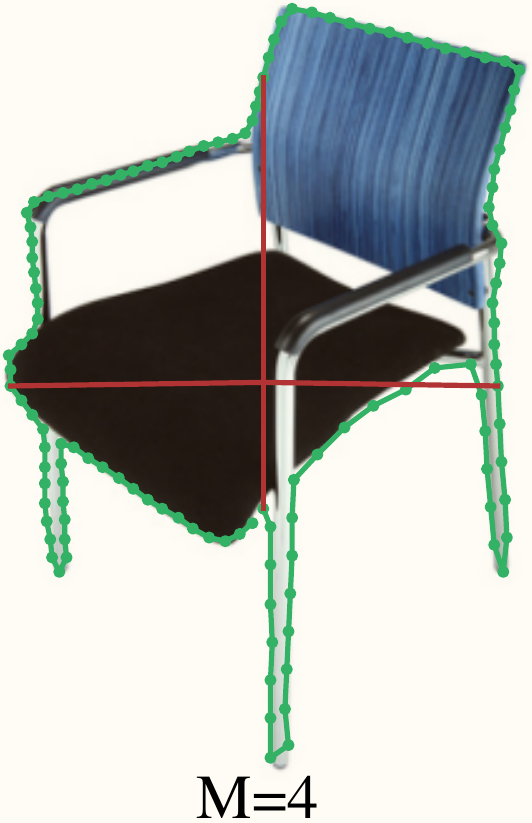}
\end{minipage}\hfill
\begin{minipage}[c]{0.25\linewidth}
\includegraphics[width=1.0\linewidth]{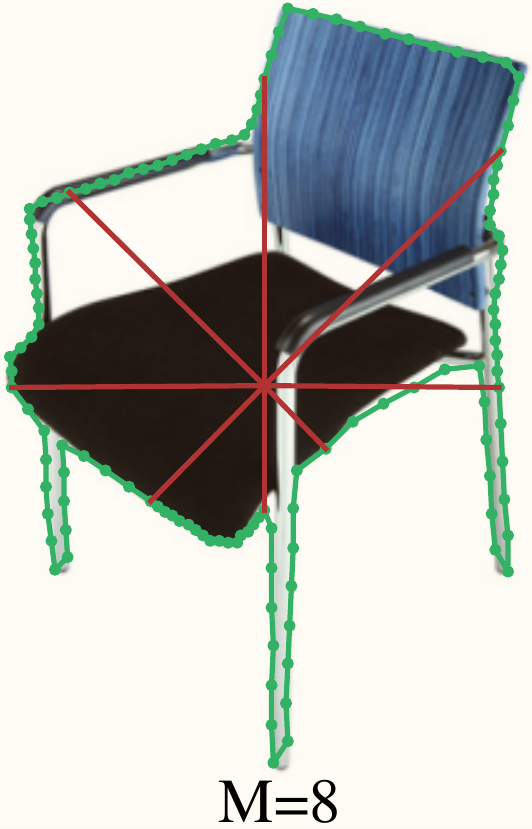}
\end{minipage}\hfill
\vspace{-3mm}
\caption{\textbf{Multi-direction alignment.} M is the number of vertices fixed in the direction with respect to the center point. When M increases, the learning difficulty of the task gradually decreases, but the unevenness of the vertex distribution also gradually increases.}\vspace{-5mm}
\label{fig:mda}
\end{figure}

\textbf{Contour-based instance segmentation methods.} Compared with the mask-based methods, the contour-based methods have an absolute advantage in speed. PolarMask \cite{polarmask} and LSNet \cite{lsnet} directly regress the coordinates of the instance vertices based on the features at the center point, and can reach a speed that is almost equivalent to that of the detector; however, the corresponding segmentation quality is quite rough. Curve GCN \cite{curvegcn}, Deep Snake \cite{deepsnake}, Point-Set Anchors \cite{pointset}, and DANCE \cite{dance} use the vertex features of the contour for the boundary regression, which greatly improves the performance. These methods first initialize the contour, and then iteratively deform the initial contour to obtain the final instance contour. However, the initial contour shapes of these methods are all manually designed, such as the ellipse of Curve GCN, the octagonal of Deep Snake, and the rectangle of Point-Set Anchors and DANCE. The huge difference between the manually designed initial contour and the ground-truth contour leads to many inappropriate vertex pairs, as shown in Figure \ref{fig:deform path}. For example, there are many intersections between the deformation paths of Curve GCN and Deep Snake at different vertices that confuse the training process. The segment-wise matching strategy proposed by DANCE slightly alleviates the above problem, but the intersections still exist. The deformation paths of Point-Set Anchors seem reasonable, but its vertex paring strategy seriously reduces the upper bound of the performance. The proposed E2EC method eliminates the unreasonable deformation path of the contour-based methods and does not reduce the upper bound of the performance.

\textbf{Other instance segmentation methods.} Dense RepPoints \cite{densereppoints} uses a discrete point set to model the instance, but the instance representation of Dense RepPoints cannot be directly converted to a mask or contour representation, and requires complex post-processing. Polytransform \cite{polytransform} combines a mask-based method and a contour-based method. Poly Transform first generates the mask representation of the instance, then converts the mask into a contour through post-processing, and finally refines the contour by a deformation module. However, Poly Transform cannot be trained end-to-end, and the speed is too slow for it to be applied to real-time scenes.
\begin{figure}[t]
\includegraphics[width=0.95\linewidth]{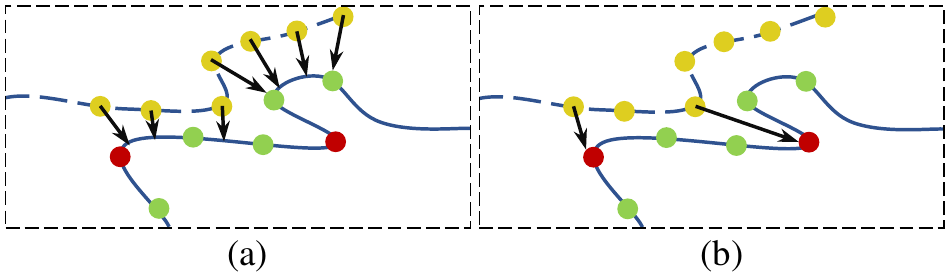}\vspace{-2mm}
\caption{\textbf{Dynamic matching loss.} The yellow points are the predicted contour vertices, the green points are the label vertices, the red points are the key label vertices, and the arrows represent the deformation path (the relationship of the pairing). (a) The first part of DML, where each prediction point is adjusted to the nearest point on the ground-truth boundary. (b) The second part of DML, where the key label point pulls the nearest prediction point toward its position.}
\label{fig:dml}\vspace{-5mm}
\end{figure}

\section{The proposed E2EC method}
\label{sec:method}

In this section, we describe the three main parts of the proposed end-to-end contour-based (E2EC) instance segmentation method, i.e., the learnable contour initialization architecture, the multi-direction alignment (MDA), and the dynamic matching loss (DML) function. The workflow of E2EC is shown in Figure \ref{fig:pipeline}. E2EC first generates a heatmap to locate the instance centers, and then learns the initial contour by regressing the initial offsets based on the center point features. The initial contour is first deformed by a global deformation module, and evolves to the coarse contour. The deformation modules \cite{deepsnake} then deform the coarse contour twice to the final contour.
\subsection{Learnable contour initialization architecture}
The learnable contour initialization architecture includes a contour initialization module and a global deformation module.

\textbf{Initial contour.} Unlike the manually designed initialization used in the existing contour-based methods, it is not necessary to specify the shape of the initial contour as this is learned by the network. Inspired by Dense RepPoints \cite{densereppoints}, the offsets of each initial contour vertex are directly regressed with respect to the center point, based on the center point features, which is denoted as $\{(\Delta x_{init}^{i},\Delta y_{init}^{i})|i=1,2,...,N\}$, where $N$ is the number of vertices of the initial contour. The initial contour vertices are computed by adding the center point coordinates and the offsets, which is denoted as $\{(x_{init}^{i},y_{init}^{i})|i=1,2,...,N\}$. Dense RepPoints regresses an unordered point set and then converts the point set into a contour or mask representation through complex post-processing. In contrast, E2EC directly regresses the contour (an ordered point set) without any requirement for post-processing. Compared with the other manually designed initial contours (e.g., ellipse or octagon), the learnable initial contour is closer to the ground-truth contour. In addition, the direction of the deformation path of the learnable initial contour is from the center point to the contour vertex (as shown in Figure \ref{fig:deform path}), guaranteeing that no unfavorable intersections between the deformation paths to impact the convergence.

\textbf{Global deformation.} It is challenging to directly regress the contour vertices with only the center point features. Meanwhile, it is also difficult to effectively deform the contour based only on the local features of a single contour vertex or several adjacent ones. The circular convolution proposed in Deep Snake uses a local aggregation mechanism to supplement the global information. However, the circular convolution operating on the local adjacent vertices needs to be repeated multiple times to aggregate the global information, and it cannot effectively correct large errors in the contour. We propose a simple but more effective global aggregation mechanism named global deformation to deform the initial contour based on both the center point features and all of the contour vertex features. As illustrated in Figure \ref{fig:global deformation}(b), the features of the $N$ initial contour vertices and the center point are first concatenated into a vector of length $(N+1)\times C$ (where $C$ is the channel number of the vertex feature). The vector is then input into the MLP module (channels of hidden layer and output layer is $N\times 2$) to obtain the offset predictions of the contour vertices (a vector of length $N\times 2$, denoted as $\{(\Delta x_{coarse}^{i},\Delta y_{coarse}^{i})|i=1,2,...,N\}$). The offsets and the initial contour coordinates are summed to obtain the adjusted coarse instance contour, which is denoted as $\{(x_{coarse}^{i},y_{coarse}^{i})|i=1,2,...,N\}$. In our experiments, we set $N=128$ and $C=64$.
\begin{figure}[t]
  \centering
   \includegraphics[width=0.95\linewidth]{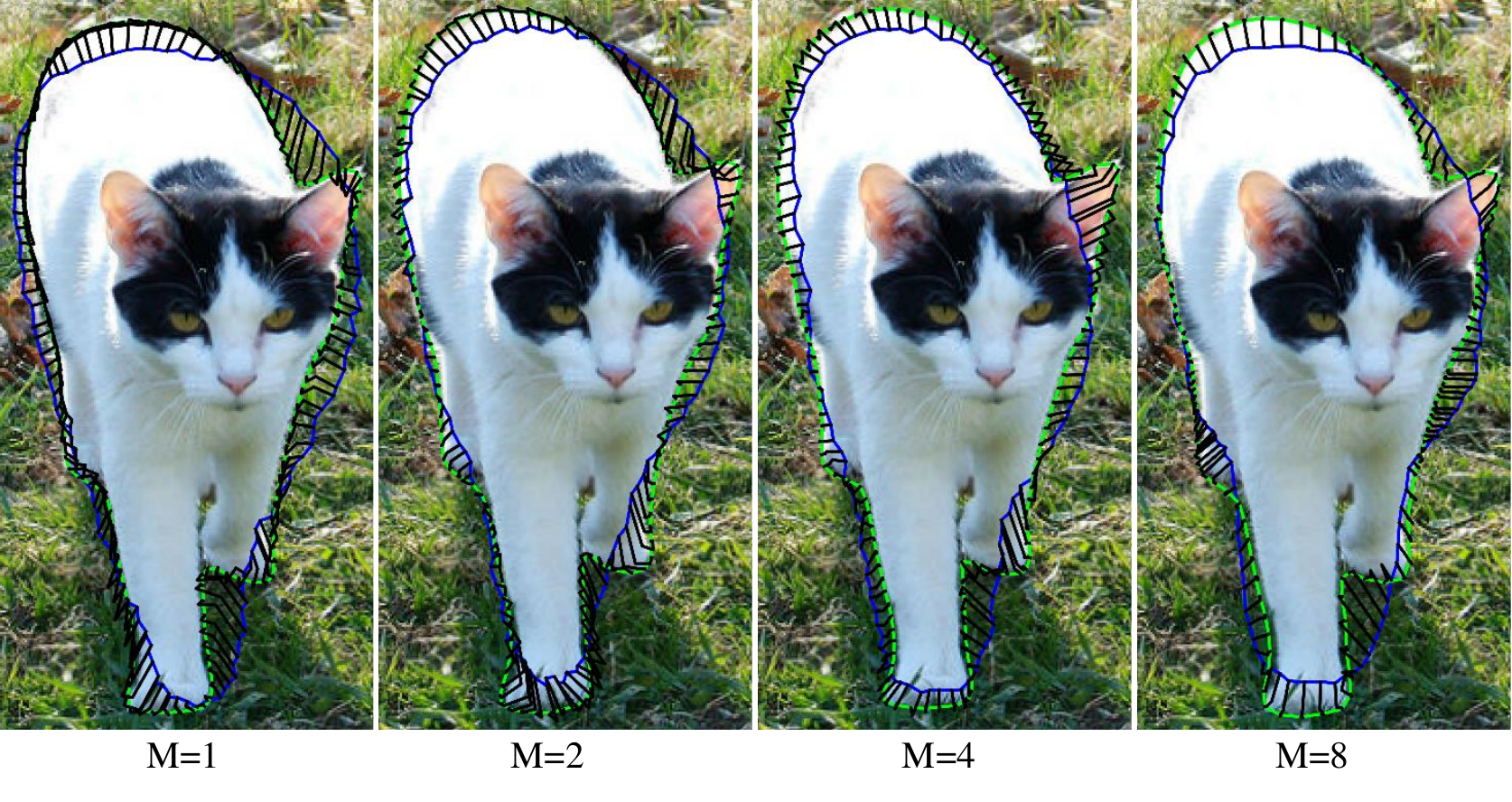}\vspace{-3mm}
   \caption{\textbf{The actual deformation paths with different numbers (M) of fixed vertices.} The blue line is the initial contour, the green line is the final contour, and the black lines are the deformation paths.}
   \label{fig:actual deform path}
   \vspace{-5mm}
\end{figure}
\subsection{Multi-direction alignment (MDA)}
Due to the challenge of the contour initialization and predicted-label vertex pairing, deviation may exist between the actual vertex deformation path and the ideal deformation path, which leads to the adjustment of some vertices tending toward the along-contour direction with a slower convergence speed, and even wrong predictions. MDA addresses this problem by fixing the direction of several selected vertices with respect to the center point, and then samples the ground truth uniformly between the fixed vertices. The sampling results for different numbers of alignment vertices are shown in Figure \ref{fig:mda}. MDA can effectively reduce the learning difficulty of the contour adjustment, without reducing the upper bound of the performance. Interestingly, PolarMask and LSNet are two extreme cases of MDA. If we suppose that the number of contour vertices is $N$ and the number of alignment vertices is $M$, when $M=N$, the strategy degenerates to PolarMask, which has the lowest learning difficulty but the lowest upper bound of the performance. When $M=0$, the strategy degenerates to LSNet, which is the most difficult case to learn, with a high upper bound of the performance. We experimentally found that $M=4$ obtains the best performance. When $M=4$, the learning difficulty is significantly reduced, but the upper bound of the performance is not reduced.

\subsection{Dynamic matching loss (DML)}
As the pre-fixed vertex pairing used in previous studies is not optimal and can cause learning difficulty, we propose DML, which dynamically adjusts the relationship of the vertex pairing to supervise output of the last deformation module \cite{deepsnake} as shown in Figure \ref{fig:pipeline}. The loss consists of two parts: 1) the predicted vertex points toward the nearest points on the label boundary, as shown in Figure \ref{fig:dml}(a), and then 2) the key label vertex pulls the nearest predicted vertex toward its position, as shown in Figure \ref{fig:dml}(b). The full details of DML are described below.

In the best case, the vertices should be adjusted to the target contour with the least cost. For each predicted vertex, it is a complicated process to dynamically find the nearest correspondence in the label contour line. Firstly, in order to simplify the calculation, the adjacent ground-truth vertices are split into 10 equal sub-segments. The problem is then transformed into discovering the nearest interpolated ground-truth contour vertex. Equation (\ref{eq1}) describes the process of matching the nearest interpolated ground-truth vertex ($gt^{ipt}$) for each predicted contour vertex by minimizing the L2 distance of predicted $i$-$th$ point and $x$-$th$ ($0<x<N+1$) label point. Equation (\ref{eq3}) is the corresponding and first component of DML. Secondly, the closest predicted vertex is dynamically matched to the key vertices (obtained by the Douglas-Peucker algorithm \cite{douglas}) of the label contour to preserve the details of the predicted contour. Equation (\ref{eq2}) describes the process of matching each key vertex to the nearest predicted vertex to best preserve the details of a boundary. Equation (\ref{eq4}) is the corresponding and second component of DML. DML is the average of the above two components, as shown in Equation (\ref{eq5}). DML can greatly improve the quality of the predicted boundary and address the over-smoothing problem found in Deep Snake and DANCE.\vspace{-2mm}
\begin{small}
\begin{equation}
x_{i}^{*}=\mathop{\arg\min_{x}}\ \|pred_{i}^{in} - gt_{x}^{ipt}\|_{2} \label{eq1}
\end{equation}\vspace{-3mm}
\begin{equation}
L_{1}(pred,gt)=\frac{1}{N} \sum_{i=1}^{N}\ \|pred^{out}_{i} - gt^{ipt}_{x_{i}^{*}}\|_{1} \label{eq3}
\end{equation}\vspace{-1mm}
\begin{equation}
y_{i}^{*}=\mathop{\arg\min_{y}}\ \|pred_{y}^{in} - gt_{i}^{key}\|_{2} \label{eq2}
\end{equation}\vspace{-3mm}
\begin{equation}
L_{2}(pred,gt)=\frac{1}{n_{key}} \sum_{i=1}^{n_{key}}\ \|pred^{out}_{y_{i}^{*}} - gt^{key}_{i}\|_{1} \label{eq4}
\end{equation}
\begin{equation}
L(pred,gt)=\frac{L_{1}(pred,gt) + L_{2}(pred,gt)}{2} \label{eq5}
\end{equation}
\end{small}\vspace{-3mm}
\section{Implementation details}
\textbf{Detector.} E2EC can be constructed based on any detector, and it is only necessary to change the output size of the bbox branch from $H\times W\times 2$ to $H\times W\times (N\times 2)$ to directly regress the initial contour with $N$ vertices. In the experiments conducted in this study, for a fair comparison with the other methods, CenterNet \cite{centernet} was used as the detector for E2EC.

\textbf{Loss function.} Smooth L1 loss is used to supervise the contour initialization branch, the global deformation branch and the first refinement deformation module. The losses are defined as:\vspace{-3mm}
\begin{small}
\begin{equation}
L_{init}=\frac{1}{N} \sum_{i=1}^{N}\ smooth\ l1(\tilde{x}_{i}^{init} - x_{i}^{gt}) \label{eq6}
\end{equation}\vspace{-3mm}
\begin{equation}
L_{coarse}=\frac{1}{N} \sum_{i=1}^{N}\ smooth\ l1(\tilde{x}_{i}^{coarse} - x_{i}^{gt}) \label{eq7}
\end{equation}
\begin{equation}
L_{iter1} = \frac{1}{N} \sum_{i=1}^{N}\ smooth\ l1(\tilde{x}_{i}^{iter1}, x_{i}^{gt}) \label{eq9}
\end{equation}
\end{small}Where $N$ is the number of contour vertices, $\tilde{x}_{i}^{init}$ is the predicted initial contour vertex, $\tilde{x}_{i}^{coarse}$ is the predicted coarse contour vertex, $x_{i}^{gt}$ is the label contour vertex, and $\tilde{x}_{i}^{iter1}$ is the contour vertex after being deformed with the first deformation module in the refinement step.

The DML function is used to supervise the last deformation module, as shown in equation (\ref{eq10}), where $\tilde{x}_{i}^{iter2}$ is the contour vertex after being deformed with the second deformation module. The loss of refinement deformation modules is then defined as equation (\ref{eq8}).
\begin{small}
\begin{equation}\vspace{-3mm}
L_{iter2} = L_{DML}(\tilde{x}_{i}^{iter2}, x_{i}^{gt}) \label{eq10}
\end{equation}\vspace{-3mm}
\begin{equation}
L_{iter} = L_{iter1} + L_{iter2} \label{eq8}
\end{equation}
\end{small}The overall loss is as follows:\vspace{-3mm}
\begin{small}
\begin{equation}\vspace{-3mm}
L_{overall} = L_{det} + \alpha L_{init} + \beta L_{coarse} + L_{iter} \label{eq11}
\end{equation}
\end{small}\vspace{-1mm}Both $\alpha$ and $\beta$ are set to 0.1. $L_{det}$ is the loss of the center point detection.

\section{Experiments}
\label{sec:experiments}

\begin{table*}[t]	
	\centering
\begin{subtable}[t]{3.2in}
	\centering
	\setlength{\tabcolsep}{1.2mm}
	\begin{footnotesize}
	\begin{tabular}{l|ccc|ccc}
	 Method & AP$^{msk}$ & AP$_{50}^{msk}$ & AP$_{70}^{msk}$ & AP$^{bdy}$ & AP$_{50}^{bdy}$ & AP$_{75}^{bdy}$ \\
	\hline
	Baseline & 54.4 & 62.1 & 48.3 & 10.8 & 35.3 & 2.6 \\ 
    +Arch & 57.5 & 64.3 & 52.2 & 17.0 & 44.9 & 9.4 \\
    +MDA & 58.8 & 65.4 & 53.9 & 18.0 & 46.8 & 10.4 \\
    +DML & 59.2 & 65.8 & 54.1 & 19.1 & 47.9 & 11.7 \\
	\end{tabular}
     \end{footnotesize}
	\caption{\textbf{Ablation experiments on the SBD dataset.} Baseline method is Deep Snake \cite{deepsnake}. Arch denotes the learnable contour initialization architecture. MDA denotes multi-direction alignment. DML denotes dynamic matching loss.}
	\label{table:component ablation}
\end{subtable}
\quad
\begin{subtable}[t]{3.2in}
	\centering
	\setlength{\tabcolsep}{1.2mm}
	\begin{footnotesize}
	\begin{tabular}{c|ccc|ccc}
	M & AP$^{msk}$ & AP$_{50}^{msk}$ & AP$_{70}^{msk}$ & AP$^{bdy}$ & AP$_{50}^{bdy}$ & AP$_{75}^{bdy}$ \\
	\hline
	1 & 57.5 & 64.3 & 52.2 & 17.0 & 44.9 & 9.4 \\ 
     2 & \textbf{58.9} & \textbf{65.8} & \textbf{54.1} & \emph{17.1} & \emph{45.8} & \emph{9.3} \\  
     4 & \emph{58.8} & \emph{65.4} & \emph{53.9} & \textbf{18.0} & \textbf{46.8} & \textbf{10.4} \\
     8 & 58.4 & 65.3 & 53.8 & 17.2 & 45.7 & 9.3 \\ 
	\end{tabular}
     \end{footnotesize}
	\caption{\textbf{Results with different alignment numbers (M).} The highest accuracy is bolded and the second-hignest accuracy is italicized.}
	\label{table:m}
\end{subtable}
\quad
\begin{subtable}[t]{3.2in}
    \centering
	\setlength{\tabcolsep}{1.2mm}
	\begin{footnotesize} 
	\begin{tabular}{c|ccc|ccc}
	Loss & AP$^{msk}$ & AP$_{50}^{msk}$ & AP$_{70}^{msk}$ & AP$^{bdy}$ & AP$_{50}^{bdy}$ & AP$_{75}^{bdy}$\\
	\hline
	smooth l1 & 58.8 & 65.4 & 53.9 &  18.0 & 46.8 & 10.4 \\
	Chamfer & 58.2 & 65.0 & 52.9 & 18.4 & 47.4 & 11.1\\
	DML & \textbf{59.2} & \textbf{65.8} & \textbf{54.1} & \textbf{19.1} & \textbf{47.9} & \textbf{11.7} \\
	\end{tabular}
     \end{footnotesize}
	\caption{\textbf{The results of the last deformation module being supervised by different loss functions.} The proposed DML outperforms smooth L1 loss and chamfer loss in terms of both the mask and boundary quality.}
	\label{table:loss}
\end{subtable}
\quad
\begin{subtable}[t]{3.2in}
	\centering
	\setlength{\tabcolsep}{1.1mm}
	\begin{footnotesize} 
	\begin{tabular}{c|ccc|ccc|c}
	Stage & AP$^{msk}$ & AP$_{50}^{msk}$ & AP$_{70}^{msk}$ & AP$^{bdy}$ & AP$_{50}^{bdy}$ & AP$_{75}^{bdy}$ & FPS\\
	\hline
	Initial& 49.8 & 61.3 & 34.4 & 1.7 & 7.5 & 0.2 & \textbf{58}\\
	Coarse& 55.7 & 64.6 & 49.5 & 8.7 & 29.4 & 2.6 & 56 \\ 
     Final& \textbf{59.2} & \textbf{65.8} & \textbf{54.1} & \textbf{19.1} &  \textbf{47.9} & \textbf{11.7} & 36 \\
	\end{tabular}
     \end{footnotesize}
	\caption{\textbf{Accuracy/speed trade-off.} Figure \ref{fig:pipeline} shows how E2EC generates the contours at these different stages.}
	\label{table:speed}
\end{subtable}

     \vspace{-3mm}
	\caption{Ablation experiments for E2EC. All models are trained on SBD train set and tested on SBD val set, using DLA-34 backbone.}
	\label{tab:ablation}
	\vspace{-2mm}
\end{table*}

\begin{table*}[t]
\begin{minipage}[c]{0.26\linewidth}
\centering
  \setlength{\tabcolsep}{0.2mm}
  \begin{footnotesize}
  \begin{tabular}{l|c|ccc}
    Method & AP$^{msk}$ & AP$_{50}^{msk}$ & AP$_{70}^{msk}$ \\
    \hline
    MNC \cite{mnc} & × & 63.5 & 41.5 \\
    FCIS \cite{fcis} & × & 65.7 & 52.1\\
    STS \cite{sts} & 29.0 & 30.0 & 6.5 \\
    ESE-20 \cite{ese} & 35.3 & 40.7 & 12.1 \\
    Deep Snake \cite{deepsnake} & 54.4 & 62.1 & 48.3 \\
    DANCE \cite{dance} & 56.2 & 63.6 & 50.4 \\
    E2EC & \textbf{59.2} & \textbf{65.8} & \textbf{54.1} \\
  \end{tabular}
  \end{footnotesize}\vspace{-4mm}
  \caption{\textbf{Results on SBD val set.}}
  \label{tab:sbd}
\end{minipage}\hfill
\begin{minipage}[c]{0.02\linewidth}
\end{minipage}\hfill
\begin{minipage}[c]{0.42\linewidth}
\centering
  \setlength{\tabcolsep}{0.8mm}
  \begin{footnotesize}
  \begin{tabular}{l|l|ccc|c}
    Method & backbone & AP & AP$_{50}$ & AP$_{75}$ & FPS\\
    \hline
    E2EC & DLA-34 & \textbf{33.8} & \textbf{52.9} & \textbf{35.9} & 30.1\\
    E2EC$^{*}$ & DLA-34 & 31.7 & 52.2 & 32.8 & \textbf{54.3}\\
    Deep Snake \cite{deepsnake} & DLA-34 & 30.3 & - & - & 27.3 \\
    PolarMask \cite{polarmask} & ResNet-101-FPN & 30.4 & 51.9 & 31.0 & 15.0 \\
    PolarMask++ \cite{polarmask++}& ResNet-50-FPN & 30.2 & 52.6 & 30.8 & 20.5\\ 
    YOLACT \cite{yolact} & ResNet-101-FPN & 29.8 & 48.5 & 31.2 & 33.5\\
  \end{tabular}
  \end{footnotesize}\vspace{-3mm}
\caption{\textbf{Results obtained on the COCO $test$-$dev$.} $^{*}$ means the two deformation modules are removed.}\label{tab:coco}
\end{minipage}\hfill
\begin{minipage}[c]{0.02\linewidth}
\end{minipage}\hfill
\begin{minipage}[c]{0.28\linewidth}
\centering
  \setlength{\tabcolsep}{0.2mm}
  \begin{footnotesize}
  \begin{tabular}{l|l|ccc}
    Datasets & Method & AP$^{bdy}$ & AP$_{50}^{bdy}$ & AP$_{75}^{bdy}$\\
    \hline
    \multirow{2}*{Kins} & Deep Snake & 30.2 & 53.0 & 31.2 \\
    ~ & E2EC & \textbf{33.3} & \textbf{54.9} & \textbf{35.6} \\
    \hline
    \multirow{2}*{SBD} & Deep Snake & 10.8 & 35.3 & 2.6 \\
    ~ & E2EC & \textbf{19.1} & \textbf{47.9} & \textbf{11.7} \\
    \hline
    \multirow{2}*{Cityscapes} & Deep Snake & 34.5 & \textbf{62.9} & 32.1 \\
    ~ & E2EC & \textbf{36.3} & 62.3 & \textbf{36.5} \\
  \end{tabular}
  \end{footnotesize}\vspace{-4mm}
  \caption{\textbf{Comparison of the boundary quality for the different datasets.}}
  \label{tab:boundary}
\end{minipage}\hfill\vspace{-3mm}
\end{table*}
\begin{table*}[t]
\begin{minipage}[c]{0.75\linewidth}
\centering
  \setlength{\tabcolsep}{0.35mm}
  \begin{footnotesize}
  \begin{tabular}{l|l|l|cc c c c c c c c c c c}
    Method & Training data & backbone & fps & AP$_{val}$ & AP & AP$_{50}$ & preson & rider & car & truck & bus & train & mcycle & bicycle\\
    \hline
    \emph{mask-based} &&&&&&&&&&&&&&\\
    SGN \cite{sgn} & Fine+coarse & VGG16 & 0.6 & 29.2 & 25.0 & 44.9 & 21.8 & 20.1 & 39.4 & 24.8 & 33.2 & 30.8 & 17.7 & 12.4 \\
    Mask R-CNN \cite{maskrcnn} & Fine & ResNet50 & 2.2 & 31.5 & 26.2 & 49.9 & 30.5 & 23.7 & 46.9 & 22.8 & 32.2 & 18.6 & 19.1& 16.0 \\
    GMIS \cite{gmis} & Fine+coarse & ResNet101 & - & - & 27.6 & 49.6 & 29.3 & 24.1 & 42.7 & 25.4 & 37.2 & 32.9 & 17.6 & 11.9 \\
    Spatial \cite{spatial} & Fine & ERFNet & 11 & - & 27.6 & 50.9 & 34.5 & 26.1 & 52.4 & 21.7	 & 31.2 & 16.4 & 20.1 & 18.9 \\
    PANet \cite{panet} & Fine & ResNet50 & $<$1 & 36.5 & 31.8 & 57.1 & 36.8 & 30.4 & 54.8 & 27.0 & 36.3 & 25.5 & 22.6 & 20.8 \\
    UPSNet \cite{upsnet} & Fine+COCO& ResNet50 & 4.4 & 37.8 & 33.0 & 59.6 & 35.9 & 27.4 & 51.8 & 31.7 & 43.0 & 31.3 & 23.7 & 19.0 \\
    \hline
    \emph{contour-based} &&&&&&&&&&&&&&\\
    Polygon RNN++ \cite{polygonrnn} & Fine & ResNet50 & - & - & 25.5 & 45.5 & 29.4 & 21.8 & 48.3 & 21.1 & 32.3 & 23.7 & 13.6 & 13.6 \\
    Deep Snake \cite{deepsnake} & Fine & DLA-34 & - & -& 28.2 & - & - & - & - & - & - & - & - & - \\
    E2EC & Fine & DLA-34 & \textbf{6.2} & 34.0 & 30.3 & 54.0 & \textbf{40.7} & \textbf{27.9} & 55.4 & 28.4 & 35.8 & 20.1 & 20.9 & 13.2 \\
    Deep Snake* \cite{deepsnake} & Fine & DLA-34 & 4.6 & 37.4 & 31.7 & 58.4 & 37.2 & 27.0 & 56.0 & 29.5 & 40.5 & 28.2 & 19.0 & 16.4 \\
    E2EC* & Fine & DLA-34 & 4.9 & \textbf{39.0} & \textbf{32.9} & \textbf{59.2} & 39.0 & 27.8 & \textbf{56.0} & \textbf{29.5} & \textbf{41.2} & \textbf{29.1} & \textbf{21.3} & \textbf{19.6} \\
  \end{tabular}\vspace{-3mm}
  \end{footnotesize}
  \caption{\textbf{Results obtained on the Cityscapes test set.} * means that the multi-component detection is used to integrate several components into a complete instance. The proposed E2EC method outperforms Deep Snake \cite{deepsnake} in all the categories.}
  \label{tab:city}
\end{minipage}\hfill
\begin{minipage}[c]{0.03\linewidth}
\end{minipage}\hfill
\begin{minipage}[c]{0.22\linewidth}\vspace{-2mm}
  \centering
  \setlength{\tabcolsep}{0.2mm}
  \begin{footnotesize}
  \begin{tabular}{l|c|cc}
    Method & AP$^{det}$ & AP$^{msk}$\\
    \hline
    \emph{mask-based} & & &\\
    MNC \cite{mnc} & 20.9 & 18.5\\
    FCIS \cite{fcis} & 25.6 & 23.5\\
    ORCNN \cite{orcnn} & 30.9 & 29.0\\
    Mask R-CNN \cite{maskrcnn} & 31.3 & 29.3\\
    Mask R-CNN* \cite{kins} & 32.7 & 31.1\\
    PANet \cite{panet} & 32.3 & 30.4\\
    PANet* \cite{kins} & 33.4 & 32.2\\
    VRS\&SP \cite{vrssp} & × & 32.1\\
    ARCNN \cite{arcnn} & × & 32.9\\
    \hline
    \emph{contour-based}& & &\\
    Deep Snake \cite{deepsnake} & 32.8 & 31.3\\
    E2EC & \textbf{36.5} & \textbf{34.0}\\
  \end{tabular}\vspace{-4mm}
  \end{footnotesize}
  \caption{\textbf{Results obtained on the KINS test set.} * denotes with ASN proposed by \cite{kins}.}
  \label{tab:kins}
\end{minipage}\hfill\vspace{-3mm}
\end{table*}

\subsection{Datasets and metrics}
\textbf{Datasets.} The KINS \cite{kins}, SBD \cite{sbd}, Cityscapes \cite{cityscapes}, and COCO \cite{coco} datasets were used in the experiments. The KINS dataset is used for amodal instance segmentation, and has seven instance classes, with 7,474 training images and 7,517 testing images. The SBD dataset has 20 instance classes and is split into 5,623 training images and 5,732 testing images. The SBD dataset is made up of 11,355 reannotated images from the PASCAL VOC \cite{voc} dataset, with instance-level boundaries. The Cityscapes dataset has eight instance classes and contains 2,975 training, 500 validation, and 1,525 testing images with high-quality annotations. The COCO dataset has eighty instance classes and contains 115k training, 5k validation, and 20k testing images. 

\textbf{Metrics.} In this paper, the mask quality is evaluated in terms of the standard AP metric. To distinguish the standard AP metric from other metrics, it is denoted as AP$^{msk}$. For all the datasets, all the settings of AP$^{msk}$ were the same as for Deep Snake.
 
The boundary quality is evaluated in terms of the boundary AP metric proposed by \cite{boundaryiou}. This is denoted as AP$^{bdy}$ and focuses on the boundary quality.

\subsection{Ablation experiments}
To quantitatively analyze the effect of each component of the proposed E2EC method and verify the design details, we conducted ablation experiments on the SBD dataset, with Deep Snake as the baseline. The network was trained end-to-end for 150 epochs with multi-scale data augmentation. The learning rate started from 1e-4 and was decayed by 0.5 at 80 and 120 epochs. The results of the ablation experiments are given in Table \ref{tab:ablation}.

\textbf{The learnable contour initialization architecture.} When the learnable contour initialization architecture is used to replace the contour initialization method of the baseline method, this yields improvements of 3.1 AP$^{msk}$, 2.2 AP$_{50}^{msk}$ and 3.9 AP$_{70}^{msk}$. For the AP$^{bdy}$ metric, which is sensitive to the boundary quality, the learnable contour initialization architecture brings improvements of 6.2 AP$^{bdy}$, 9.6 AP$_{50}^{bdy}$, and 6.8 AP$_{75}^{bdy}$. Such a huge improvement in AP fully proves the importance of a reasonable initial deformation path for the contour-based method.

\textbf{Multi-direction alignment.} When the uniform label sampling scheme used by the baseline method is replaced with the MDA strategy, this yields improvements of 1.3 AP$^{msk}$, 1.1 AP$_{50}^{msk}$ and 1.7 AP$_{70}^{msk}$ over the baseline with the learnable contour initialization architecture. For the AP$^{bdy}$ metric, also results in improvements of 1.0 AP$^{bdy}$, 1.1 AP$_{50}^{bdy}$ and 1.3 AP$_{75}^{bdy}$. It can be seen in Figure \ref{fig:actual deform path} that, as the number of alignments increases, the deviation between the actual and ideal deformation paths becomes smaller and smaller, and the actual deformation path becomes more optimal. We also conducted quantitative experiments with different alignment numbers, to explore the most appropriate number of alignment directions. The ablation results are shown in Table \ref{table:m}. The most appropriate alignment number is 4, which achieves the highest AP$^{bdy}$ and the second-highest AP$^{msk}$. As the number increases further, the difficulty of the learning decreases, but too many alignment operations also affects the upper bound of the performance and exacerbates the unevenness of the vertex distribution, as shown in Figure \ref{fig:mda}. The proposed MDA label sampling scheme makes a good balance between the learning difficulty and the upper bound of the performance.
\begin{figure*}[t]
  \centering
\begin{small}
\vspace{-1mm}\includegraphics[width=0.95\linewidth]{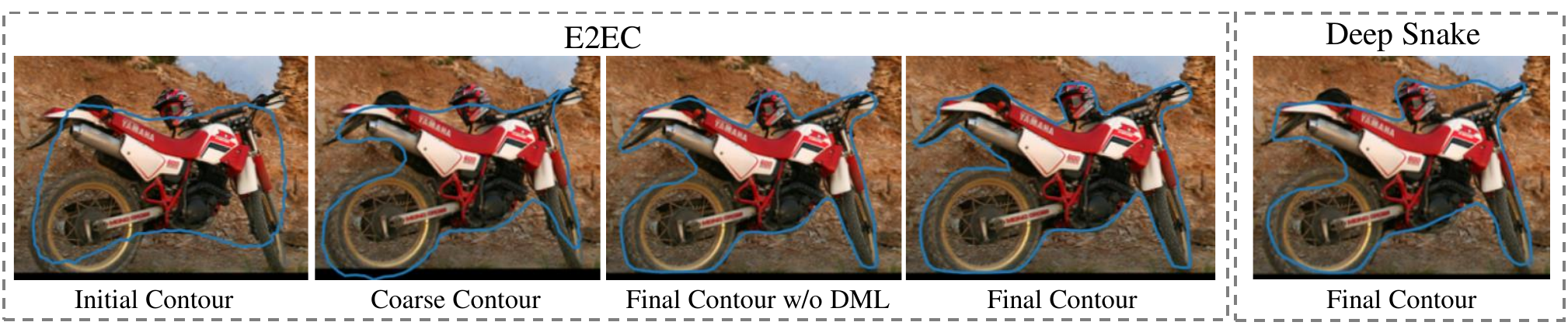}\vspace{-3mm}
   \caption{\textbf{The prediction contours at different stages.} The initial contour of E2EC fits the instance better than the octagonal initialization of Deep Snake. The quality of the coarse contour predicted by E2EC is comparable to the final contour predicted by Deep Snake. The final contour predicted by E2EC with DML perfectly outlines the motorcycle.}
   \label{fig:stage}\vspace{-4mm}
\end{small}
\end{figure*}

\begin{figure}[t]
  \centering
   \includegraphics[width=1.0\linewidth]{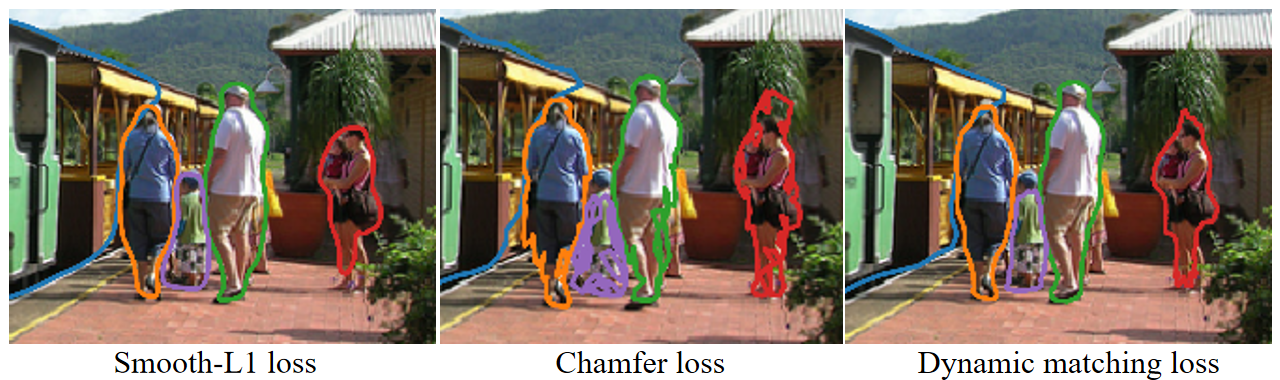}\vspace{-2mm}
   \caption{\textbf{The prediction results supervised by different loss functions.} Chamfer loss slightly improved the quality of boundary details but produced a serious jagged phenomenon (especially the red and purple instances). Dynamic matching loss heavily improved the quality of boundary details and did not cause any harm.}\vspace{-6mm}
   \label{fig:loss}
\end{figure}

\textbf{Dynamic matching loss.} Smooth L1 loss, chamfer loss \cite{chamferloss}, and DML were used to supervise the last deformation module. The results obtained with the different loss functions are shown in Table \ref{table:loss}. Replacing smooth L1 loss with chamfer loss brings about a deterioration of 0.6 AP$^{msk}$ and an improvement of 0.4 AP$^{bdy}$. Chamfer loss improves the quality of the predicted boundary, but also reduces the quality of the mask, because the order of the contour vertices has been destroyed. The proposed DML function can dynamically match all the predicted vertices with the most appropriate label vertices. DML yields improvements of 0.4 AP$^{msk}$ and 1.1 AP$^{bdy}$. The prediction results obtained with and without DML are also illustrated in Figure \ref{fig:stage}, where it can be seen that DML significantly improves the fitting degree between the predicted contour and the instance (such as the motorcycle helmet). As shown in Figure \ref{fig:loss}, the model supervised by DML predicts the best contour, without destroying the order of the vertices. 

\textbf{Speed $vs.$ accuracy.} The accuracy and inference speed of the contours at different stages are listed in Table \ref{table:speed}, and Figure \ref{fig:stage} illustrates the predicted contour at different stages. E2EC can generate an initial contour at a high speed, but the quality of the result is rough for instance segmentation. The coarse contour is obtained from the global deformation of the initial contour. The coarse contour predicted by E2EC has the same accuracy as the final output of Deep Snake. The inference speed of 56 fps on a 512×512 image far exceeds the speed of Deep Snake (33 fps). The complete E2EC achieves an inference speed of 36 fps, which is slightly higher than Deep Snake, and the final contour predicted by E2EC has a huge advantage in terms of both the mask and boundary quality. The experimental results show that E2EC predicts the coarse contour 55.5\% faster than the final contour. However, AP$^{msk}$ only reduced by 3.5. Therefore, in the case of the requirement for extremely high speed, removing the deformation components can be a good choice.
\begin{figure*}[t]
  \centering\vspace{-5mm}
   \includegraphics[width=0.98\linewidth]{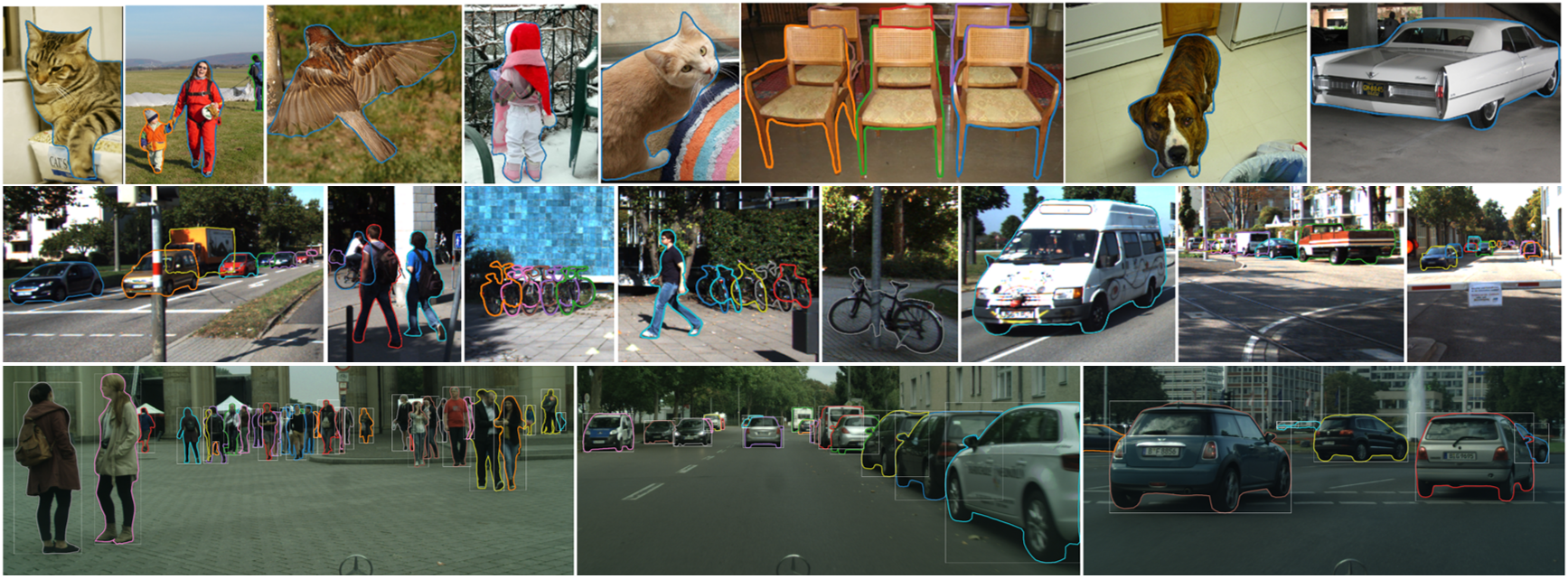}\vspace{-3mm}
   \caption{\textbf{The qualitative results obtained by E2EC.} The first row shows some examples of the prediction results for the SBD dataset, the second row shows some examples of the prediction results for the KINS dataset, and the third row shows some examples of the prediction results for the Cityscapes dataset.}\vspace{-7mm}
   \label{fig:demo}
\end{figure*}
\vspace{-3mm}\subsection{Comparison with the state-of-the-art methods}\vspace{-2mm}
\textbf{Performance on the KINS dataset.} The KINS dataset is used for amodal instance segmentation, and is annotated with inference completion information for the occluded parts of the instances. E2EC was trained for 150 epochs with the Adam optimizer, and the learning rate started from 1e-4 and was decayed by 0.5 at 80 and 120 epochs. Multi-scale training was conducted, and the models were tested at a single resolution of 768×2496.

In Table \ref{tab:kins}, the performance of the different instance segmentation methods on the KINS dataset is compared. E2EC does not use the bbox branch to generate the initial contour, so the detection result is calculated from the predicted contour. E2EC achieves a state-of-the-art performance in both the detection and segmentation tasks. E2EC achieves 36.5 AP$^{det}$, 34.0 AP$^{msk}$ and 33.3 AP$^{bdy}$ (see \ref{tab:boundary}), and outperforms Deep Snake by 3.7 AP$^{det}$, 2.7 AP$^{msk}$ and 3.1 AP$^{bdy}$. Figure \ref{fig:demo} shows some representative results of E2EC obtained on the KINS dataset.

\textbf{Performance on the SBD dataset.} For the SBD dataset, the details of the model training were the same as for the KINS dataset. Multi-scale training was conducted and the models were tested at a single resolution of 512×512.

The SBD dataset has more instance categories and more complicated instance contours than the KINS dataset, so that the advantage of the proposed E2EC method with more optimal deformation paths is more obvious. E2EC outperforms Deep Snake by 4.8 AP$^{msk}$ (see Table \ref{tab:sbd}) and 8.3 AP$^{bdy}$ (see Table \ref{tab:boundary}). The contour details predicted by E2EC are remarkable, and AP$_{75}^{msk}$ is 9.1 higher than Deep Snake. E2EC uses the lighter DLA-34 as the backbone, and it outperforms the fully convolutional instance-aware semantic segmentation (FCIS) method, which uses ResNet-101 as the backbone, by 2.0 AP$_{75}^{msk}$. Figure \ref{fig:demo} shows some examples of the results of E2EC on the SBD dataset. It can be seen that instances with complex contours, such as legs and chairs, are well segmented by the proposed method, but they cannot be well segmented by Deep Snake.

\textbf{Performance on the Cityscapes dataset.} The details of the model training for the Cityscapes dataset were the same as for the previous two datasets. It is worth mentioning that it is not necessary to train the detector separately with E2EC, but it is necessary to train the detector alone for 140 epochs and then train the whole network end-to-end for 200 epochs with Deep Snake. Multi-scale training was conducted, and the models were tested at a single resolution of 1216×2432.

Table \ref{tab:city} compares the results of the proposed E2EC method with those of the other state-of-the-art methods on the Cityscapes validation and test sets. E2EC achieves 30.3 AP$^{msk}$ on the test set, without the multi-component detection used in \cite{deepsnake}, outperforming Deep Snake by 2.1 AP$^{msk}$. With the multi-component detection strategy, the proposed method achieves 39.0 AP$^{msk}$ on the validation set and 32.9 AP$^{msk}$ on the test set, outperforming Deep Snake by 1.6 AP$^{msk}$ and 1.2 AP$^{msk}$ respectively. E2EC also outperforms the classic mask-based method of PANet by 2.5 AP$^{msk}$ on the validation set and 1.1 AP$^{msk}$ on the test set, and is almost five times faster. The performance of the proposed method is on a par with UPSNet [36]; however, the latter was trained on the additional large Common Objects in Context (COCO) dataset \cite{coco}.

Table \ref{tab:boundary} compares the boundary quality of the results obtained by the proposed E2EC method and Deep Snake. E2EC achieves 36.3 AP$^{bdy}$ and 36.5 AP$_{75}^{bdy}$ on the validation set, outperforming Deep Snake by 1.8 AP$^{bdy}$ and 4.4 AP$_{75}^{bdy}$. Some of the results obtained by E2EC are shown in Figure \ref{fig:demo}, where it can be seen that the proposed method segments the details of the instances well (such as the rearview mirror and tires), which cannot be achieved by Deep Snake.

\textbf{Performance on the COCO dataset.} For the COCO dataset, the model is trained end-to-end for 140 epochs with Adam optimizer and 20 epochs with SGD optimizer. The learning rate starts from 1e-4 and drops by half at 80 and 120 epochs, respectively. The model tested at the original image resolution without any tricks.

As shown in Table \ref{tab:coco}, E2EC achieves 33.8 AP$^{msk}$ with 30.1 fps, which outperforms Deep Snake by 3.5 AP$^{msk}$. When the deformation modules are removed, E2EC$^{*}$ achieves 31.7 AP$^{msk}$ with 54.3 fps, which is almost 2$\times$ faster than Deep Snake and outperforms Deep Snake by 1.4 AP$^{msk}$. 
\vspace{-3mm}\section{Conclusion}\vspace{-2mm}
\label{sec:conclusion}
In this paper, we have proposed an end-to-end contour-based instance segmentation method named E2EC. E2EC introduces three novel components: 1) the learnable contour initialization architecture; 2) the multi-direction alignment (MDA) label sampling scheme; and 3) the  dynamic matching loss (DML) function. E2EC greatly improves the contour extraction quality of the contour-based instance segmentation. In this study, E2EC achieved state-of-the-art results on the KINS, SBD, and Cityscapes datasets, with a beyond real-time performance. We also introduced a faster variant, where, by only retaining the learnable contour initialization architecture, the accuracy can be comparable to that of Deep Snake, and the speed is almost as fast as that of the backbone one-stage detector CenterNet. The modules proposed in E2EC could be easily applied to other contour-based instance segmentation methods. We hope that E2EC will serve as a fundamental and strong baseline for contour-based instance segmentation research.
{\small\vspace{-9mm}
\bibliographystyle{ieee_fullname}
\bibliography{ref}
}

\end{document}